\def\year{2019}\relax
\documentclass[letterpaper]{article}
\usepackage{aaai19}
\usepackage{times}
\usepackage{helvet}
\usepackage{courier}
\usepackage{url}
\usepackage{graphicx}
\usepackage{amsmath}
\usepackage{amsfonts}
\usepackage{subcaption}
\usepackage[symbol]{footmisc}

\title{Surveys Without Questions: A Reinforcement Learning Approach}
\author{
Atanu R Sinha\thanks{Co-lead authors}\thanks{Corresponding author}\textsuperscript{1},
Deepali Jain\footnotemark[1]\thanks{Performed this work at Adobe Research, now at Google AI}\textsuperscript{1},
Nikhil Sheoran\textsuperscript{1},
Sopan Khosla\textsuperscript{1},
Reshmi Sasidharan\textsuperscript{2}\\
\textsuperscript{1}{Adobe Research, India}  \textsuperscript{2}{Adobe, India}\\
atr@adobe.com, jaindeepali@google.com,\{sheoran, skhosla, rsasidha\}@adobe.com\\
}

\begin{document}
\maketitle

\begin{abstract}
The `old world' instrument, survey, remains a tool of choice for firms to obtain ratings of satisfaction and experience that customers realize while interacting online with firms. While avenues for survey have evolved from emails and links to pop-ups while browsing, the deficiencies persist. These include - reliance on ratings of very few respondents to infer about all customers' online interactions; failing to capture a customer's interactions over time since the rating is a one-time snapshot; and inability to tie back customers' ratings to specific interactions because ratings provided relate to all interactions. To overcome these deficiencies we extract proxy ratings from clickstream data, typically collected for every customer's online interactions, by developing an approach based on Reinforcement Learning (RL). We introduce a new way to interpret values generated by the value function of RL, as proxy ratings. Our approach does \textit{not} need any survey data for training. Yet, on validation against actual survey data, proxy ratings yield reasonable performance results. Additionally, we offer a new way to draw insights from values of the value function, which allow associating specific interactions to their proxy ratings. We introduce two new metrics to represent ratings - one, customer-level and the other, aggregate-level for click actions across customers. Both are defined around proportion of all pairwise, successive actions that show increase in proxy ratings. This intuitive customer-level metric enables gauging the dynamics of ratings over time and is a better predictor of purchase than customer ratings from survey. The aggregate-level metric allows pinpointing actions that help or hurt experience. In sum, proxy ratings computed unobtrusively from clickstream, for every action, for each customer, and for every session can offer interpretable and more insightful alternative to surveys. 

\end{abstract}

\section{Introduction}
With all the changes on the frontiers of the online world, one thing remains the same. An old-world instrument, survey, is still a tool of choice for firms obtaining ratings from customers about their degree of satisfaction and level of experience with online interactions on firms' websites. To obtain customer ratings, the relentless use of surveys may be surprising in the ‘new’ online world where so many rules have been rewritten. To be sure, avenues for survey have evolved from emails and links to pop-ups while browsing. That said, regardless of how surveys are conducted, firms rely upon asking questions of customers to obtain ratings of satisfaction and experience with online interactions. Unfortunately, very few customers respond. With increasingly common pop-up surveys reporting response from less than one percent of customers in our data, the sampling bias is apparent. Our proposed approach, applied on clickstream data collected by sites for \textit{every} visitor, provides proxy-ratings of online interaction experience for one hundred percent customers, without asking questions.
Henceforth, we use \textit{customer-rating} for numerical feedback obtained through surveys, and \textit{proxy-rating} for derived-feedback computed from clickstream data.
We address three major deficiencies of surveys: (i) Customer ratings obtained from very few respondents comprise the basis on which \textit{all} customers' ratings for online interactions are inferred.
The difference in ratings between the small percent who rate and the large majority of customers who do not is difficult to account and often ignored. (ii) Surveys constitute a blunt instrument since a customer's rating cannot be tied back to her specific online interactions; instead, relates to all past interactions. E.g., if a customer performs a sequence of 10 click actions at which time the survey appears and she responds, that rating cannot be tied back to specific click action(s); but, relates to the whole sequence of 10 actions. (iii) Even for the few customers whose ratings are known, a survey is a one-time, snapshot rating since the same customer may not respond more often even if surveyed.
Hence, survey ratings do not capture dynamics of online interaction experience over time. We posit that all three deficiencies can be overcome by considering customers' choice of browsing actions in a decision theoretic manner, using Reinforcement Learning (RL) \cite{sutton1988learning} and interpreting the value function in a new way.
We show that RL can facilitate drawing rich insights about individual customers and specific site interactions, which extends RL research \cite{suttonbartobookRL} in a new direction.

Consider an e-commerce platform. A customer types search words, applies filters, performs clicks to get specific product information and dives into details on few products. She may add to cart and may end up purchasing. Our premise is that since customers' ratings are best mapped to customers' behaviors, and those behaviors manifest in click actions, the clickstream data can provide useful signals of ratings. The marketing science literature informs that customers are decision oriented in their browsing behaviors \cite{moeonlinedecision}. We posit they are forward looking, learn from past and current click actions to choose future click actions, keeping in mind their eventual goals, such as search or making a purchase. This “long view” may include successive sessions where learned information from one session helps a customer decide whether or where to start browsing in the next session. 

We model this decision orientation using RL. Given a goal and a reward function, the value function of our RL model generates value of being in a state, for every state, and for every customer. States map to history of past states and click actions. Thus, we have values corresponding to every click action, for each customer, given her sequence of click actions. We interpret the values as signals of her rating at each click action, or, proxy rating at each click action. First, to test this interpretation, we validate proxy ratings of customers against actual survey ratings from the same customers. These survey ratings fit well with the goal of proxy ratings since the survey question in our data specifically seeks response on customer's website experience and not on satisfaction with product. Importantly, \textit{no survey customer ratings} are part of the model's training process. Moreover, the customer survey ratings are obtained in a natural manner, as part of pop-up surveys that the website routinely conducts (that is, not collected as an experiment). Second, making use of proxy rating values at each click action, we identify click actions that increase or decrease ratings, to identify click actions that enhance or hinder good experience. On validation against actual survey data the proxy ratings depict reasonably good performance results. On the task of action identification, we obtain insightful results about specific pages on the website that under-perform. Additionally, we test usefulness of proxy ratings on an auxiliary task of purchase prediction, which shows good performance.

We make the following contributions to the literature. One, we extend RL in a new domain of customer ratings, with focus on interpretability and on drawing rich insights from value function. Two, our approach `unobtrusively' computes proxy ratings of one hundred percent of customers. Three, proxy ratings are computed for each click action of each customer, resulting in identification of specific interactions which help or hurt customer goals. Four, proxy ratings can be obtained for each session of each customer, allowing observation of customer dynamics over time. Five, our approach does not need any survey training data. Contributions two through four address the three deficiencies of survey described at the onset. Validation of our approach against customer ratings from actual pop-up surveys is an important feature of our work. 

\section{Related Work}

Research on implicit measurement of satisfaction in the search domain focuses on metrics of dwell time, search results click, etc. to improve search outcomes. One finding is that implicit measurement correlates with explicit, question based measure ~\cite{kim2014,wang2014modeling}, lending support to our thesis. Other work in search satisfaction include a structural learning model incorporating action level dependencies using structured features ~\cite{wang2014modeling}, and studying difficulties faced while searching to obtain relevant information~\cite{odijk2015struggling}. Deviating from work in search, our problem is about decision making during interactions on online platform \cite{moeonlinedecision} to obtain ratings of interaction experience, and doing so at the granularity of every click action. A recent work in recommendation systems ~\cite{zhao2018feedbackreco_deep_rl} utilizes both explicit and implicit feedback from click-actions to learn optimal policy through trial-and-error of recommending items. Instead, our goal is to measure proxy ratings for the latent construct of experience from sequence of actions. Other research which mine clickstream data for measuring experience includes visualizations of common paths for site visitors~\cite{liu2017patterns} and inferring personas of users~\cite{zhang2016personas}. But, computation of ratings from clickstream is not addressed.

RL has an established literature with extensions in many avenues~\cite{suttonbartobookRL}. Although interpretability of machine learning models is studied~\cite{liptoninterpretability}, interpretability of value function in RL for insights about user interactions is not explored. To our knowledge, the RL literature does not examine computation of customer proxy ratings from mere clickstream data. In using clickstream data in RL, one exception is \cite{jainCX}, which measures user experience to predict purchase  through a supervised approach that uses purchase data for training. In a departure from this work, we uncover proxy ratings at each action, for each customer and perform direct evaluation against actual survey customer ratings, but, without using survey data for training.  

Recurrent Neural Networks (RNN) are used for prediction tasks from click actions ~\cite{bib17}. Problems studied include predicting sequential clicks for sponsored search~\cite{zhang2014sequential} and recognizing the sequence of tweets for purchase prediction~\cite{bib18}. We use RNN to obtain representation of states, capturing the sequential nature of actions, and use as input for the RL algorithm.

\section{Framework and Model}
\label{sec:framework}
In modeling browsing behaviors \cite{moeonlinedecision} posits that ``In addition to the decision of whether to continue searching, the consumer must also decide which item, if any, to view. At each decision point, the consumer must decide what the next item to view should be (pp.683)". We model a customer's online browsing in a decision theoretic framework because at each click action she decides whether to stop browsing, or continue browsing and if so, which click actions to select. The decisions are conditional on rewards her past actions yield and her expectation of future rewards. For example, in online shopping rewards are products she discovers, which can be good or poor, resulting in high or low reward. We recognize that a customer learns as she traverses along her sequence of actions, and that her future sequence of actions may change due to learning and is a function of goals. This process naturally fits with RL. The value generated by the value function, at each action level, to guide customer's future actions, serves as signal of her reward, which yields proxy rating. Now the formal model is presented. 

\subsection{State Representation}
In line with work in RL \cite{suttonbartobookRL}, we use a Markov process to model customers' browsing behaviors on an e-commerce site. However, drawing from Section 2.2 of ~\cite{yu2017generalized}, the Markov process is defined by using the history of past states, not only the most recent state. We augment each action with a vector containing information from history, as described next.

Consider a state space, $\mathcal{S} = \{s_1, s_2, s_3,...\}$ and a reward function $r:\mathcal{S} \rightarrow \mathbb{R}$.
At time $t$, a user in state $S_t \in \mathcal{S}$ receives a reward $r(S_t)$. The transition probability function is $\mathcal{P}(s_i, s_j) = Pr(S_{t+1}=s_j | S_t=s_i)$.
Let the sequence of click actions observed in a user's browsing journey till time $t$ be $[A_1,A_2,...A_t]$ where $A_i \in \mathcal{A} = \{a_1,a_2,...,a_{|\mathcal{A}|}\}$.
Let a vector $\vec{h}_{t-1}$ of $d$ dimensions encode all the historical information from the sequence $[A_1,A_2,...A_{t-1}]$. Then, the state at $t$ is represented as a tuple, $S_t = (\vec{h}_{t-1}, A_t)$.
Define encoding function, $g:\mathcal{S} \rightarrow \mathbb{R}^d$ such that,
	$\vec{h}_0 = \vec{0} \quad \text{and} \quad \vec{h}_t = g(\vec{h}_{t-1}, A_t) $. Here, $S_t$ comprises $\vec{h}_{t-1}$, a fixed dimension continuous vector encoding the history of actions, and $A_t$, the recent action. The fixed dimension is selected to be 150, based on limited hyper-parameter tuning (50, 100, 150, 200). Note $\vec{h}_{t}$ does not grow with $t$, being calculated using the encoding function $g$, defined above. The encoder is the hidden state of an RNN trained to predict next action.

\subsection{Reward Design} 

The choice of reward function $r$, is important in RL. We assume a website can define $r$ based on its objectives and domain knowledge. A site does not know a customer's goal. However, the site assigns reward for its objective that aligns with that of the customer's goal; e.g., making a purchase. Alternative actions of customers may be of interest to a site. A website may want to monitor the click action \textit{cart addition} for re-targeting purposes and this action can be assigned a high reward by the site. Or, if the site is interested in \textit{purchase}, this action is assigned a high reward. We assume a simple reward function for this implementation: 
    \begin{equation}
	r(S_t) = \begin{cases}	1, & \text{if $A_t$ = \textit{Purchase}}\\
	0, & \text{otherwise.}
	\end{cases}
	\end{equation}
Since purchase is of interest to a site and also a well defined customer goal we assign the purchase-action a value of 1, and for simplicity, in the absence of domain specific knowledge, assign zero values to every other action. With benefit of domain knowledge an alternative formulation could assign different rewards across actions and goals. 

\subsection{Value Iteration - TD Learning}

\begin{figure}
    \centering
    \includegraphics[width=0.95\columnwidth]{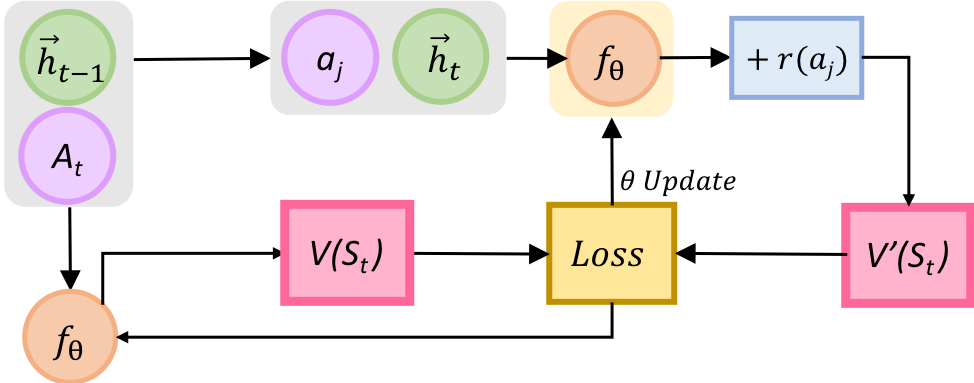}
    \caption{Illustration of Model-Free Approach}
    \label{fig:lstm-arch}
\end{figure}

We define the value of any state, $V(S_t)$ as the total expected discounted reward after $t$, under the state-transition probability distribution, $\mathcal{P}(.)$.
\begin{equation}
V(S_t) = \mathbb{E} (r(S_{t+1}) + \gamma r(S_{t+2}) + \gamma^2 r(S_{t+3}) + ...)
\end{equation}
where, $\gamma \in [0,1)$ is the discounting factor. The above expression can be written in the form of Bellman Equation as follows
\begin{equation}
	V(S_t) = \mathbb{E} (r(S_{t+1}) + \gamma V(S_{t+1}))
\label{eq:bellman}
\end{equation}

The encoder is an LSTM model trained to predict next action by using the sequence of click-actions as input. These categorical actions are embedded into a latent space of dimension 150 and fed as input to an LSTM layer that minimizes categorical cross-entropy loss on next action prediction. The hidden-representation obtained from output of the LSTM layer (at each timestep of the sequence) along with the action constitute the state representation.

To solve Equation \ref{eq:bellman}, we use an approach based on TD-learning \cite{sutton1988learning} in which the values of V(.) are estimated in a model-free fashion directly from the stream of events. In this scenario, the transition function, P(.) is not known to the model. Now consider a user transitioned from $S_t$ to $S_{t+1}$ and only current reward $r(S_{t+1})$ is observed. We make the following update to the current estimate of value function based on this observation.

\begin{equation}
\begin{split}
	& V^{'}(S_t) = r(S_{t+1}) + \gamma V(S_{t+1})\\
	& TD_t = V^{'}(S_t) - V(S_t) \\
	& V(S_t) = V(S_t) + \alpha(TD_t)
	\label{eq:td-learning}
\end{split}
\end{equation}
where, $\alpha$ is the learning rate and $V^{'}$ is the estimate of value based on the new observation. The difference in this new estimate and the current value is called the temporal-difference (TD) error. We update the current $V$ in the direction of the new estimate. After a sufficiently large number of observations the estimates converge to their fixed value. Figure \ref{fig:lstm-arch} shows the schema. 

For efficiency and generalization, we use a parameterized estimation of the value function. We define an estimation function $f_\theta$ with a set of parameters $\theta$ such that,
\begin{equation}
f_\theta(S_t) = \hat{V}(S_t) \mathrel{\hat=} V(S_t)
\end{equation}
Values of $\theta$ are randomly initialized to $\theta_0$. The optimum values are estimated through gradient descent until convergence on the TD error computed in Equation \ref{eq:td-learning} is attained.

\subsection{New Interpretation of V(.) and Metrics for Proxy Ratings}

Let us define the k-th customer's observed journey as $J^{(k)} = [A_1,A_2,...,A_m]$ and her proxy rating for action $A_t$ as $y^{(k)}_{A_t}$. The latter is equal to the computed value $V(S_t)$ at time t. Note that $m$ varies across customers. Consistent with the premise in marketing literature that satisfaction and experience ratings are interpreted as change from expectations \cite{parasuramsatisfaction}, the change of proxy ratings going from one action to the next is used as an indicator of actual ratings. We define a binary classifier for proxy ratings as follows. Given actions $A_{t-q}$ and $A_t$, we consider \textit{lag(q)} as a change in proxy ratings from $A_{t-q}$ to $A_t$. An increase in proxy ratings is assumed positive, assigned a value 1, and a decrease is assumed negative and assigned a 0. For k-th customer, define \textit{lag(q)} as:

\begin{equation}
	z^{(k)}_{A_{t-q},A_t} = \begin{cases}	1, & \text{if $y^{(k)}_{A_t}$ - $y^{(k)}_{A_{t-q}} > 0$}\\
	0, & \text{otherwise.}
	\end{cases}
\label{eq:laggedproxy}
\end{equation}

We introduce a new metric for ratings, labeled, \textit{Proportion of Good Ratings} and defined as proportion of all pairwise, successive actions (that is, $q=1$) that show increase in proxy rating values. This simple metric intuitively captures the notion of how often actions lead to better ratings. This metric is defined in two ways, $Z^{(k)}$ and $Z(a_u,a_w)$, each with its own purpose. Defined for each customer over her journey, $Z^{(k)}$ renders the proportion of her pairwise successive actions that show increase in proxy ratings. For the k-th customer, 
\begin{equation}
Z^{(k)} = \frac{1}{|J^{(k)}|-1}\sum_{t=1}^{|J^{(k)}|-1}{z^{(k)}_{A_{t-1},A_t}}
\end{equation}
For a customer performing a sequence of 20 click actions, there are 19 pairwise, successive actions. Say, 11 pairs show increase in proxy ratings. The proportion $Z^{(k)}$ is 11/19 in this example.

The second proportion, $Z(a_u,a_w)$, is defined for every pair of successive actions $(a_u,a_w)$ and represents the proportion of all instances of a pair of successive actions (note, $q=1$) that show increase in proxy ratings. 
\begin{equation}
Z(a_u,a_w) = \frac{1}{N(a_u,a_w)}\sum_{k=1}^{K}\sum_{t=1}^{|J^{(k)}-1|}{z^{(k)}_{A_{t-1},A_t}}
\end{equation}
for those $t$ where $A_{t-1}=a_u$ and $A_t = a_w$ and $N(a_u,a_w)$ denotes number of instances of successive action-pair $(a_u,a_w)$ in the data.
When a pair of successive actions occurs in 1000 instances with 350 of them showing increase in proxy ratings, the proportion $Z(a_u,a_w)$ is 350/1000. A customer can traverse the $(a_u,a_w)$ pair multiple times in a session, where each pair is a single instance. This customer contributes multiple instances to compute $Z(a_u,a_w)$. To wit, let $(a_u,a_w)$ = \textit{(ProductCategory,ProductDetail)}. It is natural for a customer to go back and forth between these two pages at different points across the length of a session. We preserve this natural phenomenon while computing $Z(a_u,a_w)$, instead of using a single average value for this customer across all instances. Use of an average value per customer loses information on variability across instances within a customer.

\section{Data}

\begin{figure}
    \centering
    \includegraphics[width=0.95\columnwidth]{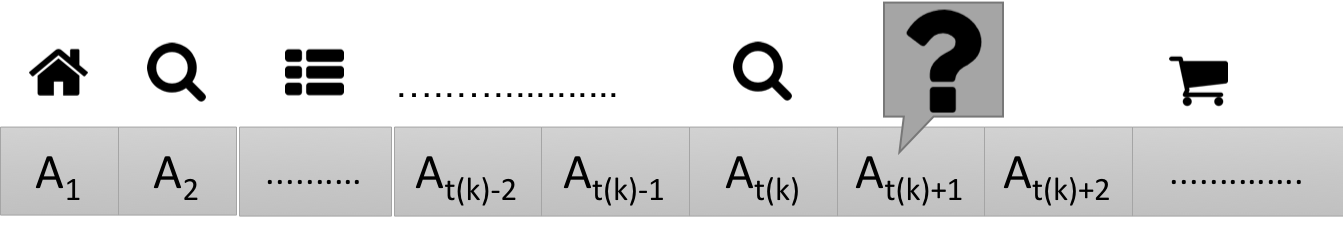}
    \caption{Sequence of actions with survey for k-th customer}
    \label{fig:action_seq}
\end{figure}

Clickstream data from the website of a consumer electronics company are used. The e-commerce site offers products in many consumer electronics product categories. For confidentiality reasons, we cannot disclose the name of the site. After filtering the data, click actions corresponding to the \textit{Laptop} category are retained. All click actions, for each customer, are stitched together chronologically into sequence of click actions. Altogether $46$ relevant click actions such as \textit{product details, search filter, add to cart} etc. are identified from the data. The set of unique actions is denoted $\mathcal{A} = \{a_1,a_2,...,a_{46}\}$. The final data are sets of sequence of actions. 

\subsubsection{Survey data} Without access to actual customer ratings survey data we cannot validate our approach. The validation is crucial to establish our thesis that values in value function work as proxy ratings. The survey appears as a pop-up, without warning, during some customers' browsing session. We do not observe in the data any systematic pattern about who gets the survey and during which part of a browsing session of a customer is the pop-up shown. The number of survey responses in the whole data is from more than 8,500 unique customers, constituting 0.7\% of all customer journeys.

The data curation pipeline is described below. Click action of \textit{Purchase} is identified in the clickstream data, whenever a purchase occurs. A customer's journey may or may not include purchase; a purchase action indicates end of journey. Most journeys do not include purchase. Also, journeys that include a purchase may qualitatively differ from those without any purchase. Based on these notions, a customer's journey can fall into one of the following categories, each of which we want represented:
\begin{enumerate}
    \item One purchase to another purchase.
    \item Starting in observation period and ending in a purchase.
    \item Beginning after a purchase in observation period and ending without purchase in observation period.
    \item No purchase throughout observation period.
\end{enumerate}

To remove outlier journeys based on length, we restrict to journeys with length in the range 10 to 210 click actions or hits. Journeys of length less than 10 click actions provide little signal into the sequence model. The distribution of hits show that journeys of less than 210 cover upwards of 96\% of all journeys. 

\subsubsection{Data Sampling} In the final dataset, we keep the ratio of purchase to no-purchase journeys as 1:2, to guard against class imbalance. Number of journeys kept in each category is as follows:
    \begin{enumerate}
        \item All journeys, roughly 7,500 in number.
        \item Around 10,000 journeys including all journeys with a survey score.
        \item Around 15,000 journeys including all journeys with a survey score.
        \item Around 20,000 journeys including all journeys with a survey score.
    \end{enumerate}

Each hit in clickstream data is mapped to an action. The final data are organized as a set of journey sequences. Each timestep in the journey contains information about the action performed, time spent, hit timestamp, customer ID and survey score (if the action corresponds to survey response). The dataset contains around 53,000 journeys from about 46,500 customers, with some customers having multiple journeys. Since we over-sample journeys which have a response to pop-up survey to obtain sufficient number of customer ratings for validation, the sampled data have a higher percentage of ratings than the 0.7\% in the whole data. The dataset is split randomly into two groups for training ($75\%$) and testing ($25\%$). Frequency distribution of a few commonly occurring actions in the training data is shown in Table \ref{tab:stats}. The training data have altogether 1,896,697 actions.  
    
\begin{table}[t]
\centering
\caption{Sample Actions and Frequencies}
\begin{tabular}[t]{|l|l|}
\hline
\textbf{Actions} & \textbf{Number} \\
\hline
\ AddToCart & 37,233 \\
\ Customize & 30,090 \\
\ Home & 119,236 \\
\ Prod.Category & 180,299 \\
\hline
\end{tabular}
\begin{tabular}[t]{|l|l|}
\hline
\textbf{Actions} & \textbf{Number} \\
\hline
\ Prod.Detail & 262,947 \\
\ Promotion & 111,648 \\
\ Search & 34,153 \\
\ ViewCart & 156,491 \\
\hline
\end{tabular}
\label{tab:stats}
\end{table}

\begin{figure*}[ht]
\includegraphics[width=\textwidth,keepaspectratio]{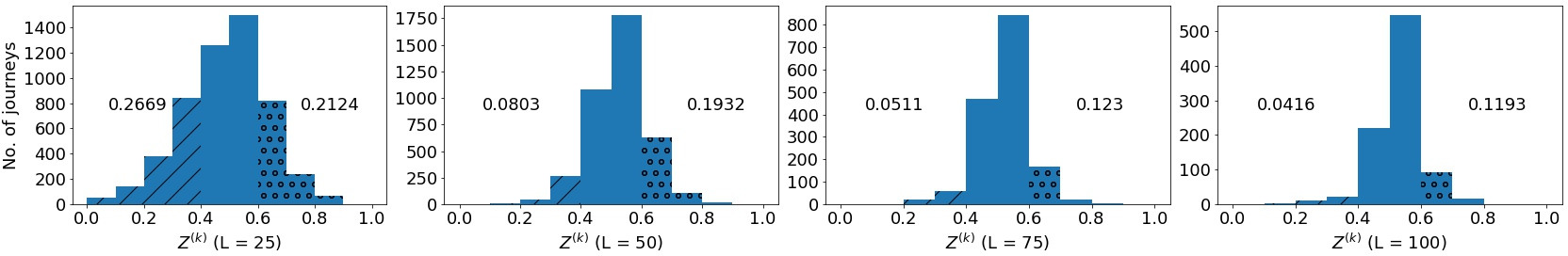}
\caption{Distribution of $Z^{(k)}$, by journey length (L)}
\label{fig:prop_good_length}
\end{figure*}

\section{Experiments and Results}
\subsection{Proportion of good proxy ratings $Z^{(k)}$}
Figure \ref{fig:prop_good_length} shows the distribution of $Z^{(k)}$ by different journey lengths. Journey lengths vary: up to ${[25, 50, 75, 100]}$. We assume that proportions in the range $(0.4,0.6)$ do not discriminate between poor and good ratings. Most of the proportions are in the range $(0.4,0.6)$ across journeys. Focusing on areas to the left and right of this range, we find that these areas are not very different from each other, for each journey length (e.g., 0.08 and 0.19 for journey length 50). We find no empirical evidence that longer journeys are associated with poorer ratings. This is not surprising because customers do not judge satisfaction merely based on reaching an end state quickly, but finding the right product for them is very important and satisfying even if the journey is longer. The finding that most journeys are in the middle $(0.4,0.6)$ is consistent with evidence that most customers are in the middle when it comes to satisfaction and experience and may not respond to surveys, thereby biasing survey ratings toward the extremes \cite{questiondesign}.

\subsection{Validation against Actual Survey}
\subsubsection{Validation Strategy}

Crucial to our validation is the pop-up survey. Consider k-th customer's sequence of click actions and pop-up survey response as shown in Figure \ref{fig:action_seq}. The survey pops up unbeknownst to the customer during her browsing session, provides an instantaneous measure, is likely to prompt top of mind response during browsing, and useful for online evaluation.
It asks \textit{``Overall, how would you rate your experience on the [company name] website today?"} on a 0-10 scale, where 10 is excellent. This question relates well to the proxy ratings we compute since these ratings are based purely on clickstream data, which are manifestations of browsing behaviors. The validation we perform is: \textit{Do proxy ratings work as leading indicators of survey responses}? The three parts to our experimental evaluation and validation are as follows.
\begin{enumerate}
\item Next action prediction to obtain representations of states capturing the history of action-sequences. 
\item Value iteration to obtain proxy ratings for each click action of each customer. 
\item Validation of proxy ratings from Part 2 against customer ratings from the pop-up survey. 
\end{enumerate}
The pop-up survey customer ratings are \textit{not} used for model training in Parts 1 and 2, and in Part 3 only used for final validation. Implementation of Parts 1 and 2 is explained in Section \ref{sec:framework}. Now we explain implementation of Part 3. From Figure \ref{fig:action_seq}, the k-th customer performs a few click actions, a pop-up appears at $A_{t(k)+1}$, after which more click actions occur. Since we want the approach to be applicable across different situations we do not want the model to know when the survey appears. This is consistent with many online pop-up surveys including the one in this data, where a customer does not know while browsing whether and when a pop-up survey may appear. Hence, the information about the survey is not treated differently from other click actions for the purpose of model training. Our premise is that a customer knows her goal, the set of click actions that are available to her, and she decides which click actions to choose to reach her goal in a decision theoretic manner. The model thus computes proxy rating values for all click actions of this customer. Also note that we do not want the model to know which customers receive pop-up survey and who do not. Thus, we compute proxy ratings for \textit{all customers} in the data based on Parts 1 and 2. Then, solely for validation in Part 3, we extract data for customers who gave actual survey ratings, since validation can only be done for those customers. Each customer's response to the survey is indicated by $A_{t(k)+1}$. Thus, for validation, we use proxy ratings up to and including $A_{t(k)}$, but do not use ratings for later actions. We do this to test whether proxy ratings up to and including $A_{t(k)}$ can predict survey responses with reasonable degree of accuracy. An occurrence of purchase soon after or much later than the pop-up survey does not impact our results. Note that the occurrence of survey-response action $A_{t(k)+1}$ in the sequence of customers' click actions vary across customer journeys. E.g., the survey may appear after 20 click actions $(t(k)=20)$ or after 100 click actions $(t(k)=100)$. To be precise, for the k-th customer, we use proxy ratings up to her action $A_{t(k)}$, but not any rating value after $A_{t(k)}$, where $t(k)$ varies across customers. Results presented are aggregated across customers by incorporating this variation. 

For each customer with a survey response, we consider \textit{lag(1)} and \textit{lag(2)} values from Equation \ref{eq:laggedproxy} for validation. With \textit{lag(1)} (substituting $q=1$ in Equation \ref{eq:laggedproxy}), if  $z^{(k)}_{A_{t-1},A_t} = 1$ we expect the actual survey rating to be \textit{good}; while if $z^{(k)}_{A_{t-1},A_t} = 0$ we expect a \textit{poor} rating. Similarly, we characterize \textit{lag(2)} ($q=2$ in Equation \ref{eq:laggedproxy}). For classification of survey score, we use a simple approach. The 0-10 scale of the pop-up survey has a natural mid-point 5, per design of the scale. We assign 0-4 as \textit{poor}, and 6-10 as \textit{good}.

\subsubsection{Validation Results}
With survey ratings and change in proxy ratings classified as good and poor, we create a confusion matrix across all respondent customers and evaluate with common metrics such as, precision, recall, accuracy, and F1. The results from the test data are presented in Table \ref{tab:results_with_survey}. The first row uses $z^{(k)}_{A_{t-1},A_t}$. The second row uses $z^{(k)}_{A_{t-2},A_t}$. We find accuracy varying between 0.63 and 0.66, recall 0.74 to 0.80, precision 0.68 to 0.69, and F1 0.71 to 0.74. We note that these numbers are reasonable, but are not high relative to domains of prediction and recommendation. That said, (i) we do not use survey data to train the model; and (ii) we posit an RL model with purchase as the only goal, while customers arrive on a site with other goals, e.g., seeking information. If customers can be grouped by goals and different goal-specific rewards assigned, we expect performance metrics to improve. Identification of customer-goals becomes an interesting research problem in its own right. In summary, we show that ratings uncovered from clickstream work as reasonable proxy for actual survey responses, with the large benefit of being obtained for every customer, and for every session or journey. 

\begin{table}
\centering
\caption{Validation on test data against Actual Survey}
\begin{tabular}{|c|c|c|c|c|}
\hline
Variation & Accuracy & Recall & Precision & F1-Score\\
\hline
$lag(1)$ & 0.6274 & 0.7408 & 0.6760 & 0.7069\\
\hline
$lag(2)$ & 0.6633 & 0.7967 & 0.6938 & 0.7417\\
\hline
\end{tabular}
\label{tab:results_with_survey}
\end{table}

\subsection{Specific Action Identification}
\subsubsection{Action Identification Strategy}

Having shown that useful proxy ratings can be obtained from clickstream, we now examine whether useful insights can be drawn for individual actions that customers perform on a site. By interpreting value function outputs as proxy ratings and introducing a new metric, we offer a systematic approach to identify actions that hinder or help toward better ratings. Since proxy ratings are computed for every click action customers perform, our approach provides insights to websites by identifying appropriate click actions which may require corrective measure. For example, if it is found that a pair of successive actions $(a_c,a_d)$ results in poor proxy score across most customers, it behooves examining this sequence for probable corrective measure such as making the click action $a_d$ less readily available when $a_c$ is clicked. We perform action identification by confining to pairs of successive actions, that is, sequences of length two. Alternatively, we can consider sequences of length greater than two. But, it is difficult to attribute a specific pair of action to a poor rating or a good rating. As an illustration, if the sequence $(a_c,a_d,a_e,a_f)$ yields poor rating, we cannot attribute which among $(a_d,a_e,a_f)$ behoove attention without additional analysis. 

For each pair of successive actions we compute the proportion of good ratings $Z(a_u,a_w)$, separately for journeys that include purchase and for those which do not include purchase. This is important because we want to avoid the potential confound of ratings with whether customers end up with a purchase. Thus, for each of purchase and no-purchase conditions, with 46 unique actions in our data, we have a 46x46 matrix, or, 2116 cells. However, since all action pairs are not observed in the data (e.g., \textit{(Home, Add To Cart)}), empirically we have 1861 populated cells. Note that the successive actions in a pair can be identical; e.g., \textit{(ProductDetail, ProductDetail)}. Most action-pairs are not popularly traversed by users. A site can choose to focus on action-pairs traversed above a threshold value.

\begin{table}
\centering
\caption{Actions impacting experience}
\begin{tabular}{|c|c|c|c|}
\hline
Source & Target & Purch. & $Z(a_u,a_w)$ \\
\hline
\hline
Customize & ProductCategory & No & $0.58\pm 0.010$\\
\hline
Customize & ProductCategory & Yes & $0.44\pm 0.009$ \\
\hline
\hline
Customize & ProductDetail & No & $0.52\pm 0.014$ \\
\hline
Customize & ProductDetail & Yes & $0.77\pm 0.013$\\
\hline
\hline
Home & ProductCategory & No & $0.68\pm 0.011$\\
\hline
Home & ProductCategory & Yes & $0.63\pm 0.013$ \\
\hline
\hline
Home & ProductDetail & No & $0.22\pm 0.012$\\
\hline
Home & ProductDetail & Yes & $0.16\pm 0.009$ \\
\hline
\end{tabular}
\label{table:actionidentification}
\end{table}

\begin{figure*}[ht]
\includegraphics[width=\textwidth,keepaspectratio]{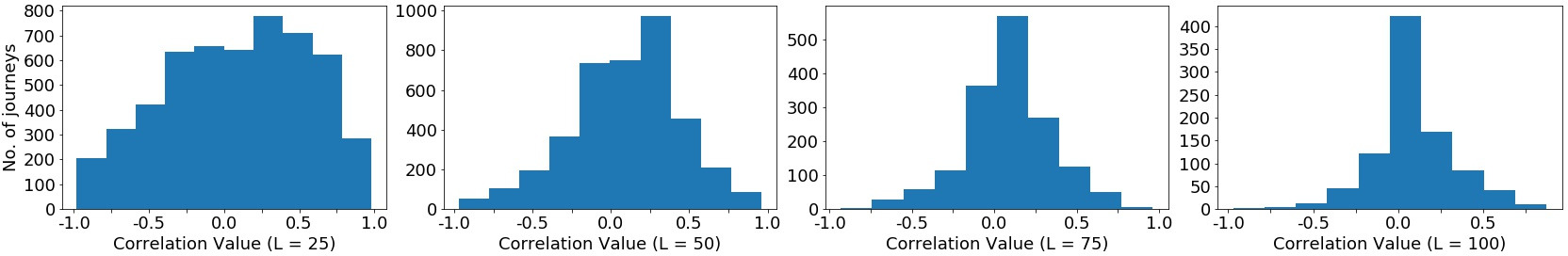}
\caption{Distribution of Correlations between $y^{(k)}_{A_t}$ and probability of purchase, by journey length (L)}
\label{fig:xv_pop}
\end{figure*}

\begin{figure*}[ht]
\includegraphics[width=\textwidth,keepaspectratio]{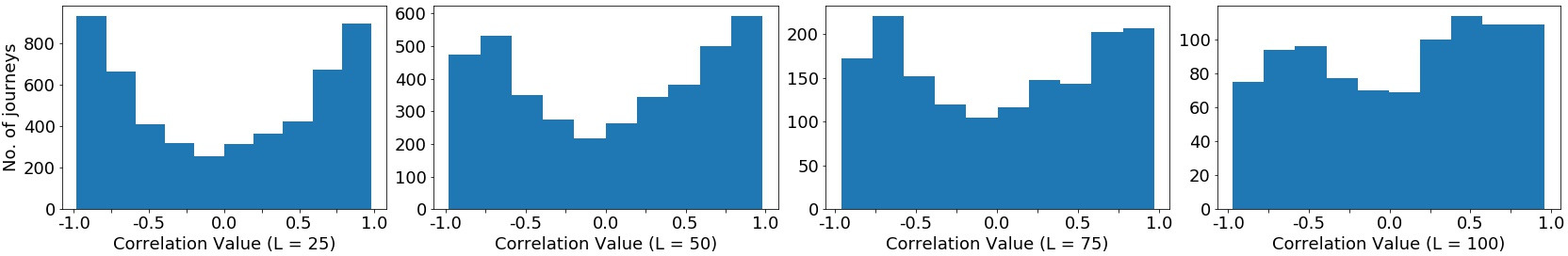}
\caption{Distribution of Correlations between $Z^{(k)}$ and probability of purchase, by journey length (L)}
\label{fig:pos_pop}
\end{figure*}

\subsubsection{Action Identification Results}
Table \ref{table:actionidentification} conveys results for selected sample pairs as a way to illustrate the types of interpretation and insight these proxy ratings can provide. The results are presented in pairs of rows, corresponding to (\textit{No,Yes}) in \textit{Purchase} column, for every pair of successive actions (\textit{Source, Target}). Reviewing the last column $Z(a_u,a_w)$ we note that if a proportion is close to 0.50 it is not discernible for action identification, since good and poor proportions are similar. We make two types of comparison: (a) for each successive pair of actions (\textit{Source, Target}), compare proportions for those who purchase and those who do not; (b) for each type of customer groups - non-purchasers and purchasers, compare proportions across action-pairs. Considering (a), the proportions do not show systematic differences between purchase and no-purchase groups, although these differences are statistically significant at 5\%. In three action-pairs purchasers show a lower score, while higher in one pair (\textit{Customize, ProductDetail}). Across all 1861 pairs in the data, we find no systematic differences between these two groups, providing support that our approach does not necessarily associate proportion of good ratings with purchase. This is consistent with the literature that satisfaction and experience are not simply about purchase \cite{lemon} and also shows that our approach to proxy ratings is not biased toward purchase. Focusing on (b), first 
consider action-pairs (\textit{Customize, ProductCategory}) and (\textit{Customize, ProductDetail}) for \textit{Purchase = Yes}. Moving from \textit{Customize} to \textit{ProductDetail} generates significantly more good proxy ratings than moving from \textit{Customize} to \textit{ProductCategory} (mean values 0.77 versus 0.44). This can be interpreted using the well-known concept of purchase funnel \cite{purchasefunnel}, where transition along a funnel moves consumers from one stage to the next; e.g., \textit{Home} to \textit{ProductCategory} to \textit{Customize} to \textit{ProductDetail}. If the transition moves to a stage further ahead by skipping a stage, that may go against a smooth transition. To wit, \textit{Customize} is a more detail-oriented activity. A move from \textit{Customize} to \textit{ProductCategory} is a regressive step, while moving to \textit{ProductDetail} is a progressive step. Thus, the latter is associated with higher good proxy rating. Next, consider action-pairs (\textit{Home, ProductCategory}) and (\textit{Home, ProductDetail}). Within each of these pairs of actions, the no-purchase and purchase conditions show mean values that are numerically close, namely, (0.68, 0.63) and (0.22, 0.16), respectively. Across action-pairs, for both no-purchase and purchase conditions, the values are different. Going from \textit{Home} directly to \textit{ProductDetail}, by skipping \textit{ProductCategory}, is being interpreted as less fulfilling (0.22, 0.16), because often customers backtrack to  \textit{ProductCategory} and re-start the funnel. The (\textit{Home, ProductCategory}) is a natural progression and favorable scores (0.68, 0.63) support that interpretation. Thus, from (b), the differences show that proxy scores can be interpreted to yield useful insights which are associated with customer behavior in a marketing funnel.

\section{Discussion and Conclusion}

We validate proxy ratings against actual survey scores and identify relevant actions using proportions. As an auxiliary goal, we now address purchase prediction since it is of common interest. In the following paragraphs we discuss two purchase prediction tasks - one, predicting purchase at each timestep of click action for every customer, and two, predicting whether a journey ends up with a purchase.

The benchmark model for task one is an LSTM trained in a supervised manner to predict probability of purchase at every timestep. For each timestep, our model yields proxy rating as well as proportion of good ratings. For every customer-journey, we compute correlation across time steps, between probability of purchase and proxy rating $y^{(k)}_{A_t}$. Figure \ref{fig:xv_pop} shows distributions of correlations by lengths of journey. The correlations are centered around zero as uni-modal distributions. With journeys of length up to 25, the variance is large, but decreases with longer journeys. We also compute, for every customer-journey, correlation across time steps, between probability of purchase and proportion of good ratings $Z^{(k)}$. Figure \ref{fig:pos_pop} shows these correlations form bi-modal distributions and depict different patterns from that of Figure \ref{fig:xv_pop}. This suggests that the proportion $Z^{(k)}$ is a better discriminator than proxy rating $y^{(k)}_{A_t}$ in relating to probability of purchase at each timestep. Additionally, our enumerated metrics are distinct from purchase probabilities at every timestep and capture information about customer interactions that purchase probability models do not. 

Coming to task two, we know which journeys end up in purchase. We enumerate $Z^{(k)}$ for every customer, across the person's whole journey. Then we use $Z^{(k)}$ to predict whether or not purchase occurs at the end of the journey. We obtain an AUC = 0.73. If we use actual customer survey scores to predict purchase, an AUC = 0.51 is obtained, which is no better than random prediction. Note that for the latter we confine only to customers who provide survey ratings, most of whom do not purchase. The metric $Z^{(k)}$ appears useful to predict eventual purchase in a journey. Last but not the least, yet another value-add from our model is that proxy ratings can be computed every time a customer browses on a site and thus her satisfaction and experience over sessions and journeys can be ascertained. This unobtrusive measure can provide early warning through downward trend in her proxy ratings. This cannot be done using surveys.

In conclusion, we show that proxy ratings and derived metrics of proportion are interpretable and insightful and can serve as reasonably good alternatives to surveys. With much advancement in online presence of firms, surveys have remained prevalent despite well-known deficiencies.
We offer a way out of this situation through a reinforcement learning based extrication of proxy ratings from easily available clickstream data.
This research takes RL in a new direction to better understand interactions in customer data and brings RL a step closer to realizing its potential in the online firm-customer interaction domain.


\bibliographystyle{aaai}
\end{document}